\documentclass{article}
\usepackage{ijcai19}

\usepackage{times}
\usepackage{soul}
\usepackage{url}
\usepackage[hidelinks]{hyperref}
\usepackage[utf8]{inputenc}
\usepackage[small]{caption}
\usepackage{graphicx}
\usepackage{amsmath}
\usepackage{booktabs}
\usepackage{graphicx}
\usepackage{pict2e}
\usepackage{amssymb}
\usepackage{amsthm} 
\usepackage{subfigure}
\usepackage{graphics}
\usepackage{wrapfig,lipsum,booktabs}
\usepackage{etoolbox}
\usepackage{caption}
\usepackage{graphicx}
\usepackage{pict2e}
\usepackage{amssymb}
\usepackage{amsthm} 
\usepackage{graphicx}
\usepackage{epsfig}
\usepackage{epstopdf}

\title{Mesh Learning Using Persistent Homology on the Laplacian Eigenfunctions}

\author{
Yunhao Zhang$^1$\and
Haowen Liu$^1$\and
Paul Rosen$^{2}$\And
Mustafa Hajij$^1$\\
\affiliations
$^1$Ohio State University\\
$^2$University of South Florida\\
\emails
\{zhang.7581, liu.7040\}@buckeyemail.osu.edu,
prosen@usf.edu,
hajij.1@osu.edu
}

\begin{document}

\maketitle

\begin{abstract}
We use persistent homology along with the eigenfunctions of the Laplacian to study similarity amongst triangulated 2-manifolds. Our method relies on studying the lower-star filtration induced by the eigenfunctions of the Laplacian. This gives us a shape descriptor that inherits the rich information encoded in the eigenfunctions of the Laplacian. Moreover, the similarity between these descriptors can be easily computed using tools that are readily available in Topological Data Analysis. We provide experiments to illustrate the effectiveness of the proposed method.
\end{abstract}

\section{Introduction}
\label{sec:intro}

Shape similarly is a critical problem is computer vision, geometric data processing and computer graphics. Multiple attempts have been made to quantify the similarity among 3D shapes \cite{biasotti2014quantifying,tam2013registration,skopal2011nonmetric}. Several challenges rise up when trying to construct an effective and efficient similarity measure including the complexity of the data, the potential noise in the data and the variation in the structure.


While the Laplacian eigenfunctions have been utilized in the literature of geometric processing to extract shape descriptors \cite{sun2009concise,zhang2007survey}, most of the eigenfunction-based descriptors require extensive processing to obtain an effective descriptor. Furthermore, the comparison between such descriptors  requires designing a specialized similarity measure that adds to overhead computational time \cite{van2011survey}.  

The eigenfunctions of the Laplacian store important information about the geometry of the underlying manifold \cite{reuter2006laplace,rustamov2007laplace}. Moreover, spaces that have similar structures also tend to have similar sets of eigenfunctions \cite{levy2006laplace}. From this perspective it is natural to utilize the eigenfunctions to measure the similarity between a collection of 3D shapes. The difficulty usually lies in finding the correspondence between two given manifolds \cite{van2011survey}. More specifically, when manifold is discretized this correspondence might not even exist due to the difference between the cardinalities of the two vertex sets. Instead sub part correspondence might be considered, which is also a difficult problem \cite{biasotti2006sub}. 

In recent years the interplay between machine learning and Topological Data Analysis (TDA) has witnessed many developments with the better understanding of two tools in TDA: \textit{Persistent Homology} (PH) \cite{EdelsbrunnerLetscherZomorodian2002} and the construction of \textit{Mapper} \cite{singh2007topological}. These TDA tools have been shown to be a powerful tool for shape classification and recognition~\cite{kusano2016persistence,reininghaus2015stable}, data summary~\cite{hajij2018graph,choudhury2012topological}, topological signatures of data~\cite{bonis2016persistence}, graph understanding~\cite{hajij2018visual}, among others.

In this paper we utilize persistent homology to extract the information encoded in the eigenfunctions of the Laplacian to obtain a topological mesh signature that can be used to measure the similarity among triangulated manifolds. Our proposed method has multiple advantages. On one hand, the method proposed here avoids the correspondence problem all together. Our approach relies on extracting the topological information encoded into the \textit{lower-star filtration} (see Section \ref{Ph on TM} for the definition) of the eigenfunctions of the Laplacian and storing the resulting finger print in the a structure called the \textit{persistence diagram} \cite{EdelsbrunnerLetscherZomorodian2002}. This ultimately allows for an effective comparison between two manifolds by comparing between the persistence diagrams that are induced by the eigenfunctions of the Laplacian. 

Using the persistence diagram to compare between two metric spaces has been previously applied to meshes \cite{chazal2009gromov}. However, the metric-based method in \cite{chazal2009gromov} has two major limitations. Firstly,  finding the distance function on large meshes is computationally expensive and usually requires utilizing a sampling technique, which might affect the quality of the final persistence diagram. Secondly, in order to obtain a strong descriptor from the persistence diagram induced by the distance matrix, one usually needs the information encoded in higher order persistence diagrams, which are expensive to compute. 

Our method avoids these two limitations. On one hand, our method computes the persistence diagram using the lower-star filtration of one or a few eigenfunctions of the Laplacian. In fact we show that utilizing a single eigenfunction yields a persistence diagram that has more classification power than the metric-based approach in \cite{chazal2009gromov}.  On the other hand, our method only requires the $0$-order persistence diagram, which is very efficient to compute. We demonstrate our results by showing the effectiveness of our descriptor on standard datasets. See Section \ref{results} for more details.

\section{Persistence Homology on Triangulated Meshes}
\label{Ph on TM}
The mesh topological signature that we propose here utilizes a particular filtration that is induced by a scalar function defined on a mesh $M$. Our work is mainly aimed at studying triangulated meshes. However, we will state our definitions in terms of simplicial complexes. The reason for this is that most techniques introduced in this article are applicable beyond meshes, and we will provide more details in this regard towards the conclusion.



Let $K$ be a simplicial complex. Let $S$ be an ordered sequence $\sigma_1,\cdots,\sigma_n $ of all simplices in the complex $K$, such that for simplex $\sigma \in K $ every face of $\sigma$ appears before $\sigma$ in $S$. Then $S$ induces a nested sequence of subcomplexes called a \textit{filtration}:
\begin{equation}
\label{filter}
 \phi=K_0  \subset K_1 \subset ... \subset K_n = K   
\end{equation}
such that $K_i = \cup_{j\leq i } \sigma_j$ is the subcomplex obtained from first $i$ simplicies $\sigma_1,\cdots,\sigma_i $ of $S$.  Given a filtration as in equation \ref{filter}, one may apply the homology functor on it to obtain a sequence of homology groups connected by homomorphism maps induced by the inclusions:
\begin{equation}
 \mathcal{F}(K): H_d(K_0) \longrightarrow H_d(K_1) \longrightarrow ... \longrightarrow H_d(K_n) 
\end{equation}
A $d$-homology class $\alpha \in H_d(K_i)$ is said to be \textit{born} at time $i$ if it appears for the first time as a homology class in $H_d(K_i)$. A $d$-class $\alpha$ \textit{dies} at time $j$ if it is trivial $H_d(K_j)$ but not trivial in $H_d(K_{j-1})$. The \textit{persistence} of $\alpha$ is defined to be $j-i$. \textit{ Persistent homology} captures the birth and death events in a given filtration and summarizes them in a multi-set structure called the \textit{persistence diagram} $P^d(K)$ \cite{EdelsbrunnerLetscherZomorodian2002}. Specifically, the $d$-persistence diagram of the a filtration $\mathcal{F}(K)$ is a collection of pairs $(i,j)$ in the plane, where each $(i,j)$ indicates a $d$-homology class that is created at time $i$ in the filtration $\mathcal{F}(K)$ and dies entering time $j$.  A persistence diagram can be represented equivalently by \textit{persistence barcodes}~\cite{Ghrist2008}. Specifically every point $(i,j)$ in the persistence diagram can be represented by a \emph{bar} that starts at time $u$ and ends at time $v$.     

Persistence homology tracks the evolution of homology classes as this element moves though the homomorphism from left to right. More specifically, Persistent homology can be defined given any filtration, such as equation \ref{filter}. For our purpose we are given a piece-wise linear function $f:|K|\longrightarrow \mathbb{R}$ defined on the vertices of $K$. We assume that the function $f$ has different values on different vertices of $K$. Any such a function induces a filtration called the \textit{lower-star} filtration. We define this filtration next.

Let $v\in V(K)$ be a vertex of $K$. The \textit{star} of $v$, denoted as $St(v)$, is the set of all simplices in $K$ that contain $v$ as a vertex. When we are given a piece-wise linear function $f$ defined on $K$, we can also define the lower star of $v$. Namely, the lower star of a vertex $v \in V(K)$ as $LowSt(v)=\{w \in St(v)| f(w)\leq f(v)\}$. 

Let $V=\{v_1,\cdots,v_n\}$ be the set of vertices of $K$ sorted in non-decreasing order of their $f$-values. Let $K_i:= \cup_{j\leq i } LowSt(v_j) $. The lower-star filtration is the filtration is defined to be 
\begin{equation}
\label{filter2}
\mathcal{F}_f(K):  \phi=K_0  \subset K_1 \subset ... \subset K_n = K   
\end{equation}

The lower-star filtration reflects the topology of the function $f$ in the sense that the persistence homology induced by the filtration, equation \ref{filter2}, is identical to the persistent homology of the sublevel sets of the function $f$. We denote $P^d_f(K)$ to be the $d$-persistence diagram induced by the lower-star filtration $\mathcal{F}_f(K)$. In our work, the lower-star filtration is the main tool to extract the signature from a given space.

Here we focus on triangulated meshes, and we only compute the $0$-persistence diagram on those meshes using the filtration induced by the lower-star filtration of the eigenfunctions of the Laplacian of these meshes. Such persistence diagrams can be efficiently computed using the \textit{union-find} data structure.

\subsection{The Lower-Star Filtration Induced by the Eigenfunctions of the Laplacian.} 
Let $M$ be an triangulated manifold.
The matrix $L$ is self-adjoint and positive semi-definite. It has an \emph{orthonormal eigensystem} \mbox{$(\lambda_{n},\phi_{n})_{n=0}^{+\infty}$}, \mbox{$L\phi_{n}=\lambda_{n}\phi_{n}$}, with \mbox{$0=\lambda_{0}\leq\lambda_{n}\leq\lambda_{n+1}$}, in $C(G)$. The eigenvectors of the Laplacian $L$ form a rich family of scalar functions defined on $G$ that have been utilized extensively in shape understanding and shape comparison \cite{LafonLee2006,reuter2006laplace,Levy2006}. The eigenfunctions of the Laplacian has also been used in graph understand~\cite{ShumanNarangFrossard2013}, segmentation~\cite{ReuterBiasottiGiorgi2009}, and spectral clustering~\cite{NgJordanWeiss2002}.

The reasons for extracting the information of the eigenfunctions of the Laplacian using the lower-star filtration can be summarized in the following points: 

\begin{itemize}
    \item The eigenfunctions of the Laplacian provide canonical scalar functions that depend only on the intrinsic geometric properties of the mesh. In other words, they have all desirable properties of eigenfunctions of the Laplacian---being invariant under certain deformation and robustness to noise and structure variation---will be inherited by the persistence diagram induced by the lower-star filtration of these functions. 
    \item  The eigenfunctions of the Laplacian store rich information about the geometry of the underlying manifold and the lower-star filtration provides the means to extract this information and store it in the structure of the persistence diagram. This structure provides a ranking for features extracted from the eigenfunctions via the notion of persistence.      
\end{itemize}


Ordering the eigenvectors of $L$ by the increasing value of their corresponding eigenvalues, we use the first $k$-eigenvectors that correspond to the smallest nonzero $k$ eigenvalues of $L$. These vectors contain low frequency information about the underlying manifold, and they usually retain the shape of complex meshes. In particular, we found that the first non-trivial eigenfunction of the Laplacian to be very effective for our purpose. This eigenfunction, called the \textit{Fiedler vector}~\cite{fiedler1973algebraic,fiedler1975property}, has many applications in graph theory as well as in computer graphics~\cite{isenburg2005streaming, mullen2008spectral}. Moreover, this vector has multiple interesting features. For instance, the maximum and the minimum of the Fielder vector tend to occur at points in the dataset with maximum geodesic distance \cite{ChungSeoAdluru2011} allowing its values to spread from one end of the graph following its ``shape'' to the other end.

\begin{figure}[!b]
	\centering
		\includegraphics[height=92pt]{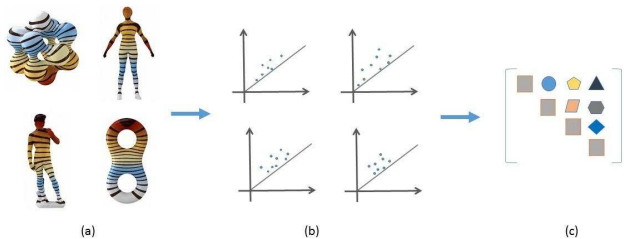}
	\caption{An illustration of the pipeline. (a) We compute one of the eigenfunctions of the Laplacian on the meshes that we want to compare. (b) The lower-star filtrations of the meshes with respect to these scalar functions are computed and their persistence diagrams are extracted. (c) A pairwise comparison between the persistence diagrams is performed using the bottleneck or Wasserstein distance.}
	\label{fig:header}
\end{figure}

\begin{figure*}[!ht]
	\centering
	\includegraphics[height=180pt]{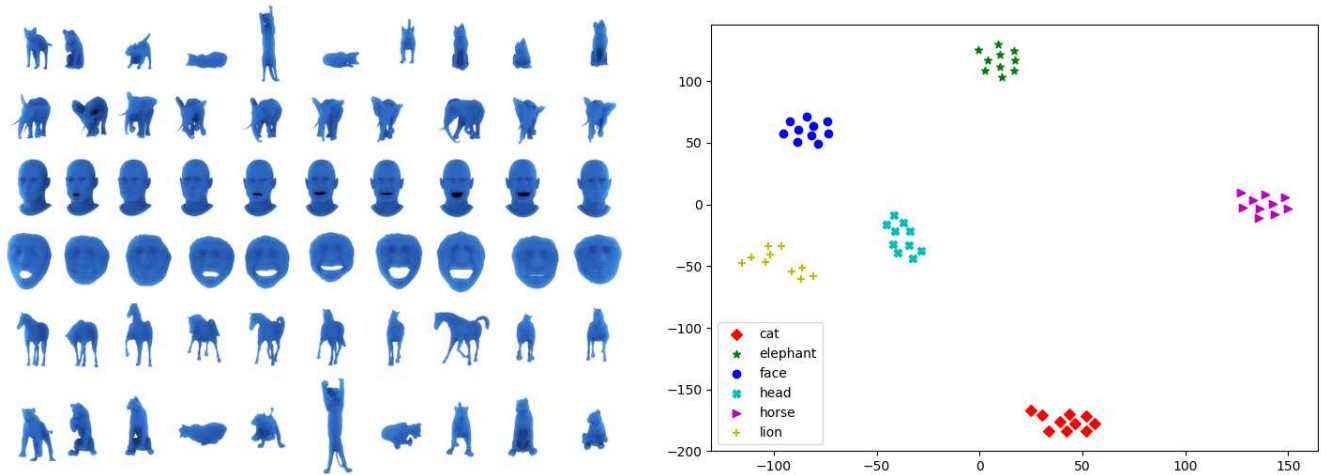}
	\caption{On the left we show the data set that we used in our experiments. The data set consists of $60$ triangulated meshes that are divided into $6$ categories, which are shown in the figure on the right. We compute the the Fielder's vector for each mesh in this data set and then compute the $0$-persistence diagram associated with the lower-star filtration of this vector. The bottleneck distance between these diagrams is calculated, and the figure on the right shows the 2d t-SNE projection obtained using the final distance matrix. Notice how our method provides distinct clusters on this data.}
	\label{fig:tnse projection}
\end{figure*} 

\subsection{Comparing Between Two Persistence Diagrams}
\label{distance}

We can quantify the structural differences persistence diagrams by using the bottleneck and Wasserstein distances. 


Let $\eta$ be a bijection between two persistence diagrams $X$ and $Y$. The bottleneck distance between $X$ and $Y$~\cite{EdelsbrunnerHarer2010} is defined as 
\begin{equation}
W_{\infty}(X,Y) = \inf_{\eta: X \rightarrow Y} \sup_{x \in X} \left\lVert x-\eta(x) \right\rVert_\infty.
\end{equation}
The Wasserstein distance between the diagrams is $X$ and $Y$ is defined as
\begin{equation}
W_q(X,Y) = \displaystyle \left[ \inf_{\eta:X \rightarrow Y}  \Sigma_{x\in X} \left\lVert x-\eta(x) \right\rVert^q_\infty \right]^{1/q}, 
\end{equation}
for any positive real number $q$; in our setting, $q=2$. Both of these distances requires the persistence diagrams to have the same cardinalities. For this reason we allow infinitely replication of points along the diagonal $y = x$ to a given persistence diagram.

\section{Method and Results}
\label{results}
Given the above setup, our method can be summarized as follows. First we compute a certain eigenfunction of the Laplacian on a given mesh dataset. In our case we used the Fielder's vector. We then compute the $0$-persistence diagrams of the lower-star filtration induced by the chosen eigenfunction. Once we have the persistence diagrams of the meshes, the distances between these diagrams can be computed using one of the distances we defined in Section \ref{distance}. 
See Figure \ref{fig:header} for a summary of the method.


To validate the effectiveness of the topological descriptor proposed here, we test it using a publicly available data from \cite{sumner2004deformation}. The data set consists of $60$ meshes that are divided into 6 categories: cat, elephant, face, head, horse and lion. Each category contains exactly ten triangulated meshes.

On this dataset, we computing the distance matrix between the persistence diagram of the lower-star filtration of the induced Fielder's vectors using bottleneck distance. To assess the final results, we compute the 2d t-SNE projection \cite{vanDerMaaten2008} of final distance matrix. The result is reported in Figure \ref{fig:tnse projection}.

Note how this topological descriptor provides a distinct clusters for the underlying data set. Furthermore, the results shown here shows that the proposed descriptor has a better classification power than the one proposed in \cite{chazal2009gromov}.

\section{Further Directions and Conclusion}

The experimentation results are only shown with respect to Fielder's vector. In theory any eigenfunction of the Laplacian can be used in a similar manner, as illustrated above. Combining the signatures obtained from multiple eigenfunction potentially provides even a stronger descriptor. We plan to pursue this direction in the extension of this work.

The construction that we introduced here on triangulated meshes can be easily extended to study similarity between other types of objects. Namely any domain where the definition of the Laplacian is applicable, such as points clouds and graphs. We plan to investigate these directions in the future.


\bibliographystyle{plain}
\bibliography{refs}
\end{document}